\documentclass{article}

\usepackage[preprint]{neurips_2026}

\usepackage[utf8]{inputenc} 
\usepackage[T1]{fontenc}    
\usepackage{hyperref}       
\usepackage{url}            
\usepackage{booktabs}       
\usepackage{amsmath}        
\usepackage{amsfonts}       
\usepackage{nicefrac}       
\usepackage{microtype}      
\usepackage{xcolor}         
\usepackage{graphicx}       
\usepackage{multirow}       
\usepackage{xspace}         
\usepackage{pifont}         
\usepackage{array}

\newcommand{\modelname}{Tango3D\xspace}

\title{\modelname: Towards Alignment for Global and Local 2D-3D Correspondence}

\author{%
 Zebin He$^{1,2}$\thanks{Work done during internship at Tencent Hunyuan.} \quad Mingxin Yang$^{2}$ \quad Shuhui Yang$^2$ \quad Hanxiao Sun$^{1,2}$ \\ \textbf{Xintong Han$^2$ \quad Chunchao Guo$^2$\thanks{Corresponding Author.} \quad Wenhan Luo$^{1\dag}$}
 \\\\
 $^1$HKUST \quad $^2$Tencent Hunyuan \\
}

\begin{document}

\maketitle

\begin{figure}[h]
  \centering
  \includegraphics[width=\linewidth]{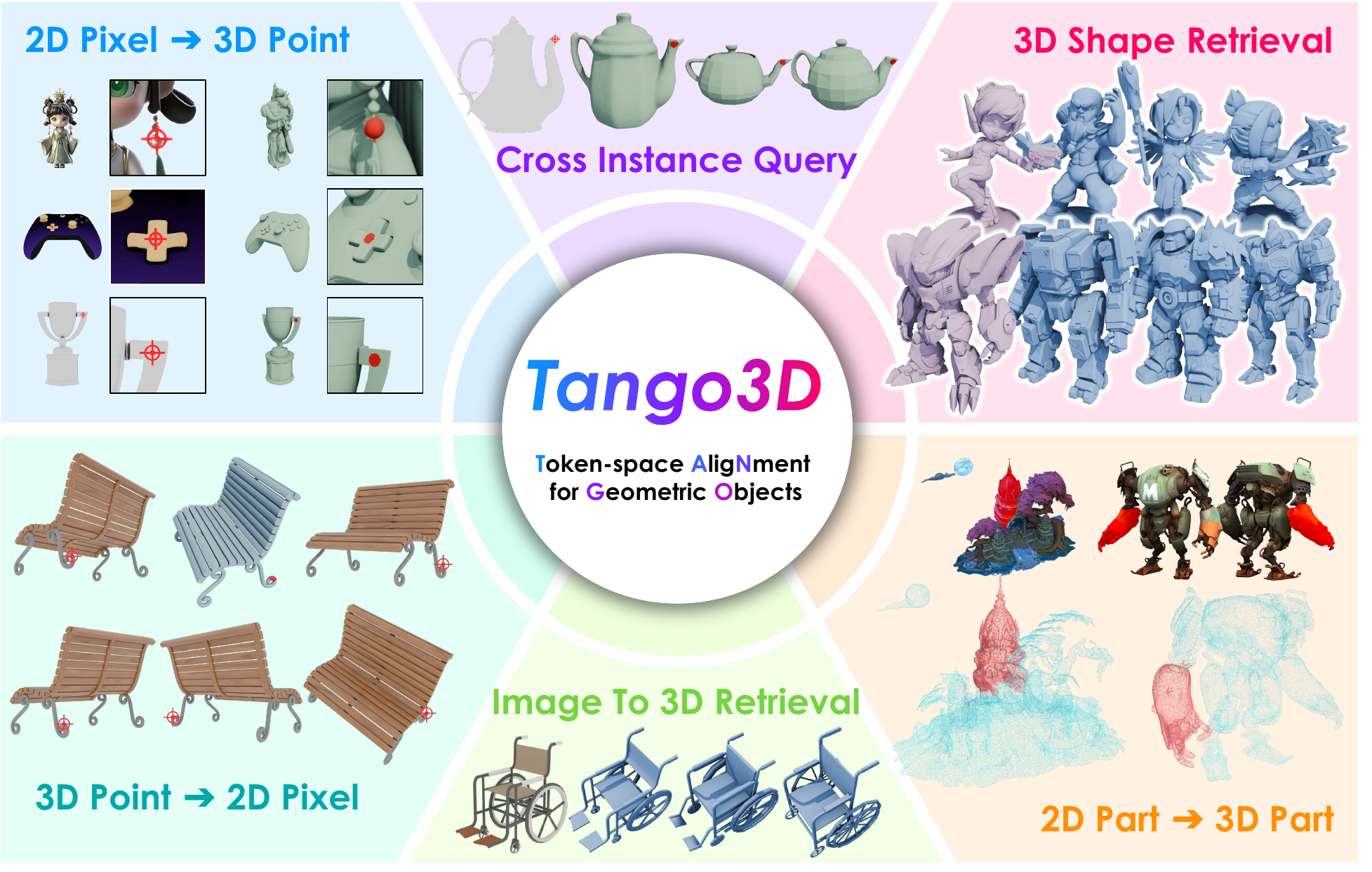}
  \caption{Our framework \modelname unifies fine-grained correspondence and global retrieval, supporting diverse cross-modal and intra-modal matching tasks from pixel-level grounding to instance-level retrieval.}
  \label{fig:teaser}
\end{figure}

\begin{abstract}
Existing 3D foundation models typically align point clouds to frozen vision-language spaces like CLIP, which achieve strong cross-modal retrieval by compressing 3D shape into a global vector. However, this global-only alignment cannot establish fine-grained pixel-to-point correspondence.
To solve this, we present \modelname, a foundation model that unifies dense correspondence and global retrieval. We use a geometry-aware 2D visual backbone and a pretrained 3D VAE to encode images into 2D patches and point clouds into 3D tokens. These are mapped into a single shared space to achieve both local pixel-to-point alignment and global semantic alignment.
To stabilize the joint learning of dense and global objectives, we introduce a three-stage progressive training strategy. Experiments show our model successfully achieves object-level pixel-to-point alignment while maintaining competitive global retrieval, a joint capability not offered by existing 3D foundation models. By establishing a fine-grained alignment feature space, \modelname injects rich semantics into purely geometric 3D tokens, paving the way for a wide range of dense 3D downstream tasks.
\end{abstract}

\section{Introduction}

Aligning 3D point clouds with 2D visual foundations has become the standard approach for 3D representation learning. By training 3D encoders against frozen vision-language spaces like CLIP~\cite{pmlr-v139-radford21a}, recent 3D foundation models have achieved remarkable success. This alignment paradigm successfully injects open-vocabulary semantics into 3D representations, establishing a strong foundation for zero-shot classification and shape retrieval.

However, this paradigm suffers from a fundamental limitation: the alignment is exclusively global. Existing methods compress an entire 3D shape into a single feature vector. They typically use purely semantic spaces like CLIP as alignment anchors, which are designed for semantic richness but inherently lack spatial structure. Consequently, while sufficient for retrieval, this compression discards the spatial details needed for pixel-to-point correspondence. Such dense alignment is critical for fine-grained tasks like 3D part segmentation. Although scene-level methods~\cite{NEURIPS2024_e7e506bc, Li_2025_CVPR} have explored dense matching, object-level 2D-3D dense alignment remains an open problem.

To solve this, we present \modelname, a foundation model that unifies dense correspondence and global retrieval. Instead of relying on purely semantic spaces, we use the geometry-aware 2D visual backbone VGGT~\cite{wang2025vggt} as our alignment anchor to provide the necessary spatial structure. Furthermore, rather than compressing 3D shapes into a single vector, we utilize a 3D VAE to encode point clouds into 3D tokens. By establishing a shared token space, 2D feature patches and 3D tokens can interact directly. This design naturally supports fine-grained pixel-to-point alignment while preserving strong global semantic capabilities.

However, jointly optimizing both dense correspondence and global retrieval presents a significant challenge. To stabilize this learning process, we introduce a three-stage progressive training strategy. This strategy guides the model from establishing robust fine-grained local alignments to integrating global semantic representations, ensuring that the dense and global objectives mutually enhance rather than conflict with each other. 
Experiments demonstrate that \modelname successfully achieves object-level pixel-to-point correspondence while maintaining competitive global retrieval performance. Our approach also enables zero-shot semantic matching at both the object and point levels. By bridging the gap between 2D and 3D modalities, our approach lays the foundation for exploring robust and unified representations, which could potentially benefit future complex 3D downstream tasks.

In summary, our main contributions are:

\textbf{Local alignment.} We propose \modelname, a foundation model that achieves object-level pixel-to-point dense correspondence by mapping 2D patches and 3D tokens into a shared space. The learned space supports bidirectional queries between pixels and points, cross-instance semantic matching at the point level, and 3D-to-3D correspondence without 2D input.

\textbf{Global alignment.} We unify dense correspondence and instance-level retrieval within a single framework. The aggregated global descriptors support both image-to-shape and shape-to-shape retrieval, achieving competitive performance against leading global-only baselines.

\textbf{Progressive training.} We introduce a three-stage progressive strategy to stabilize the joint optimization. It first optimizes geometric correspondence, then integrates global semantics, and finally refines both at higher resolution, leading to a more coordinated optimization of the two objectives.

\section{Related Work}

\textbf{Multi-modal 3D representation learning.} 
CLIP~\cite{pmlr-v139-radford21a} learns a joint image and text embedding space, and subsequent works~\cite{Zhai_2023_ICCV, tschannen2025siglip} further improve efficiency. 
Recent advances in 3D representation learning have rapidly evolved from designing robust unimodal architectures~\cite{NEURIPS2022_d78ece66, wu2024ptv3, qian2022pointnext} to multi-modal alignment frameworks. Early approaches adopt lightweight 3D encoders~\cite{Qi_2017_CVPR, NIPS2017_d8bf84be, Yu_2022_CVPR, pang2022masked} or depth encoders~\cite {Huang_2023_ICCV}, and align their outputs to the frozen CLIP embedding space.
ULIP~\cite{Xue_2023_CVPR} and ULIP-2~\cite{Xue_2024_CVPR}
construct image, language, and 3D triplets and train a Point-BERT backbone.
OpenShape~\cite{NEURIPS2023_8c7304e7} focuses on data curation by ensembling multiple 3D datasets with filtered and enriched text descriptions.
All of these methods use a frozen CLIP ViT (typically ViT-bigG-14) as the 2D encoder and only train the 3D branch.
Uni3D~\cite{ICLR2024_cc2c03ce} directly reuses a 2D ViT as the 3D encoder and achieves strong zero-shot classification.
Beyond instance-level alignment,
MRD~\cite{wang2024multi}
distills relational structure from the pretrained CLIP space, preserving both intra-modal and cross-modal similarity distributions.
MixCon3D~\cite{Gao_2024_CVPR}
combines multi-view CLIP image features with point-cloud features into a holistic embedding for text alignment.
OpenView~\cite{Zhou_2025_CVPR} further introduces a cross-modal fusion encoder that jointly processes point clouds and multi-view images.

Despite their success, these methods are limited to global-level alignment and cannot establish dense pixel-to-point correspondences, as both the single-vector compression and the purely semantic CLIP anchor discard the spatial structure required for fine-grained geometric grounding.

\textbf{2D-3D dense correspondence.} 
Traditional works establish 2D-3D correspondences via multi-stage SfM pipelines~\cite{schoenberger2016sfm, snavely2006photo, agarwal2011building, frahm2010building, wu2013towards, liu2025robust}.
More recent efforts instead learn these correspondences end-to-end in a differentiable manner~\cite{zhou2017unsupervised, ummenhofer2017demon, tang2018ba, wei2020deepsfm, wang2021deep, teed2018deepv2d, teed2021droid, brachmann2024acezero, smith24flowmap:, wang24vggsfm:}. 
DUSt3R~\cite{dust3r_cvpr24} predicts dense per-pixel 3D coordinates from image pairs, MASt3R~\cite{mast3r_eccv24} adds explicit local descriptors trained with a matching loss to ground correspondences in 3D, Fast3R~\cite{Yang_2025_CVPR} generalizes the pairwise design to process many views in a single forward pass, and Sem-MASt3R~\cite{Tenore_2025_ICCV} injects DINOv2 semantics to disambiguate textureless regions. 
While achieving impressive 2D-to-3D correspondences, these methods target scene-level reconstruction and do not address cross-modal alignment between image and point clouds. Other scene-level efforts further explore pixel-to-point matching via contrastive distillation or implicit cross-attention~\cite{NEURIPS2024_e7e506bc, Li_2025_CVPR}. In contrast, our work targets object-level 2D-3D alignment across modalities, and simultaneously maintains a global retrieval capability.

\section{Method}

\label{sec:method}
\begin{figure}[t]
  \centering
  \includegraphics[width=\linewidth]{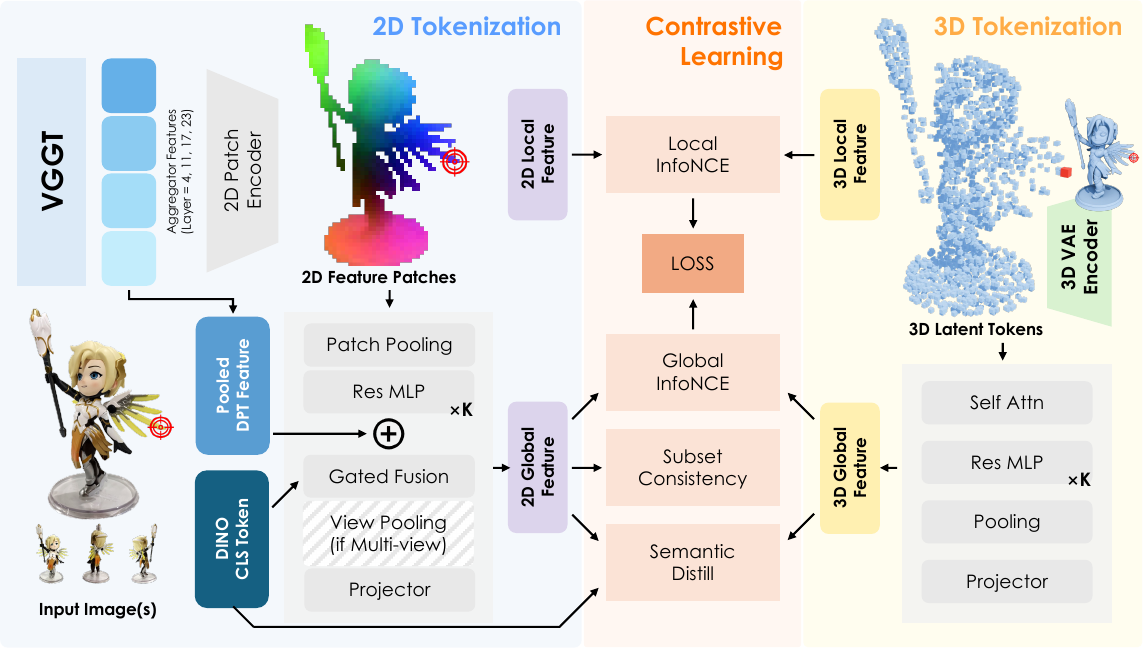}
  \caption{\textbf{Overview of \modelname.} We project 2D patch features from a frozen VGGT backbone and 3D latent tokens from a trainable VAE into a shared embedding space, jointly optimized by a local branch for pixel-to-point correspondence and a global branch for instance-level retrieval.}
  \label{fig:overview}
  \vspace{-5mm}
\end{figure}

As illustrated in Fig.~\ref{fig:overview}, given image(s) $\mathcal{I} = \{I_s\}_{s=1}^{S}$ ($S \geq 1$) and a surface point cloud $\mathcal{P} \in \mathbb{R}^{N \times 6}$, our framework learns a unified embedding space that simultaneously supports fine-grained 2D-3D correspondence and instance-level cross-modal retrieval. To achieve this, we propose a dual-branch architecture built upon a shared token space: a local branch for geometry-grounded dense alignment, and a global branch for object-level semantic aggregation. 

\subsection{2D and 3D tokenization}
\label{subsec:tokenization}

\begin{table}[t]
\centering
\caption{\textbf{Pixel-to-point local correspondence.} We evaluate dense geometric alignment by retrieving the nearest 3D tokens for randomly sampled 2D foreground pixels. LocAcc@k is reported under both single- and multi-view configurations.}
\label{tab:local}
\small
\begin{tabular}{l cc ccccc}
\toprule
 & \multicolumn{2}{c}{Baselines} & \multicolumn{5}{c}{\textbf{\modelname(ours)}} \\
\cmidrule(lr){2-3} \cmidrule(lr){4-8}
Input views & VGGT~\cite{wang2025vggt}@1 & VGGT-OBJ~\cite{chang2026reconviagen}@1 & @1 & @2 & @3 & @5 & @10 \\
\midrule
S=1, random       & {--} & {--} & 77.43 & 80.24 & 82.02 & 84.23 & 87.14 \\
S=4, random      & {--} & {--} & 77.37 & 80.26 & 82.07 & 84.37 & 87.38 \\
S=4, ortho       & 72.04 & 72.16 & \textbf{76.37} & 79.49 & 81.42 & 83.85 & 87.00 \\
\bottomrule
\end{tabular}
\vspace{-2mm}
\end{table}

Our framework leverages two pretrained models to extract geometry-aware representations. 
For the 2D and 3D modality, we employ VGGT~\cite{wang2025vggt} and the encoder of the Hunyuan3D 2.1 variational autoencoder~\cite{hunyuan3d2025hunyuan3d} respectively
To extract 2D features, each view $I_s$ is processed by the frozen VGGT backbone and its DPT refinement head in a single forward pass, producing four intermediate feature maps and one fused DPT feature map. These feature maps are resized to the patch grid of the current training stage and concatenated into a 3072-dimensional feature vector $x_{s,u,v} \in \mathbb{R}^{3072}$ per grid position. The fused DPT map is also globally averaged to yield a 256-dimensional per-view context $c_s$. During Stage~I and Stage~II, VGGT operates at $518 \times 518$ input resolution, producing a $37 \times 37$ patch grid; in Stage~III, the resolution is increased to $1022 \times 1022$, giving a $73 \times 73$ grid. 
To construct geometry-aware 2D queries for local supervision, we sample each rendered position map at the same grid centers. We aggregate valid patches across views and retain at most 512 queries via farthest-point sampling. For each retained query, we extract its 3D coordinate $ q_m $, its corresponding feature vector $ x_m $, and record its source-view index $ \pi(m) $.

For the 3D modality, the point cloud is processed by a trainable 3D VAE encoder, yielding 1024 latent tokens $z_n \in \mathbb{R}^{D_{\mathrm{vae}}}$ and their spatial centers $p_n \in \mathbb{R}^3$. We use the encoder features before posterior sampling, which retain richer geometric detail than the bottleneck latent.

Finally, both modalities are mapped into a shared 1024-dimensional space via residual MLP encoders:
\[
h_m^{2\mathrm{D}} = f_{2\mathrm{D}}(x_m,\; c_{\pi(m)}), \qquad h_n^{3\mathrm{D}} = f_{3\mathrm{D}}(z_n),
\]
where the 2D encoder injects the per-view context. This shared token space serves as the unified foundation for the entire framework, directly supervised by the local branch and aggregated by the global branch.

\subsection{Local branch}
\label{subsec:local_branch}
We project the shared tokens $h_m^{2\mathrm{D}}$ and $h_n^{3\mathrm{D}}$ into modality-specific local descriptors via linear heads and $\ell_2$-normalization. For each 2D query $q_m$, we identify its positive 3D token via nearest-neighbor assignment in the geometric space: $n^\star(m) = \arg\min_n \|q_m - p_n\|_2$. 

Because the projected 2D queries may not perfectly align with the discrete 3D token centers, we apply a Gaussian confidence weight $w_m = \exp(-\|q_m - p_{n^\star}\|_2^2 / 2\sigma^2)$ to each query and exclude 3D tokens within a spatial distance $\delta$ of $q_m$ from the negative set. We optimize these descriptors using a bidirectional InfoNCE contrastive loss:
\[
\mathcal{L}_{\mathrm{local}} = \frac{1}{2}\bigl(\mathcal{L}_{2\mathrm{D}\to 3\mathrm{D}} + \mathcal{L}_{3\mathrm{D}\to 2\mathrm{D}}\bigr),
\]
where each query's contribution to the loss is weighted by $w_m$. Instead of relying solely on the forward matching, the reverse direction explicitly aggregates all 2D queries assigned to the same 3D token. This bidirectional formulation prevents the network from collapsing into a many-to-one mapping. During later stages, the loss is augmented with top-$k$ hard negatives~\cite{NEURIPS2023_8c7304e7, Zhou_2025_CVPR} in each direction, where both the number of hard negatives $k$ and their loss weight are increased across stages to progressively sharpen discriminability.

\subsection{Global branch}
\label{subsec:global_branch}

\begin{figure}[t]
    \centering
    \includegraphics[width=\linewidth]{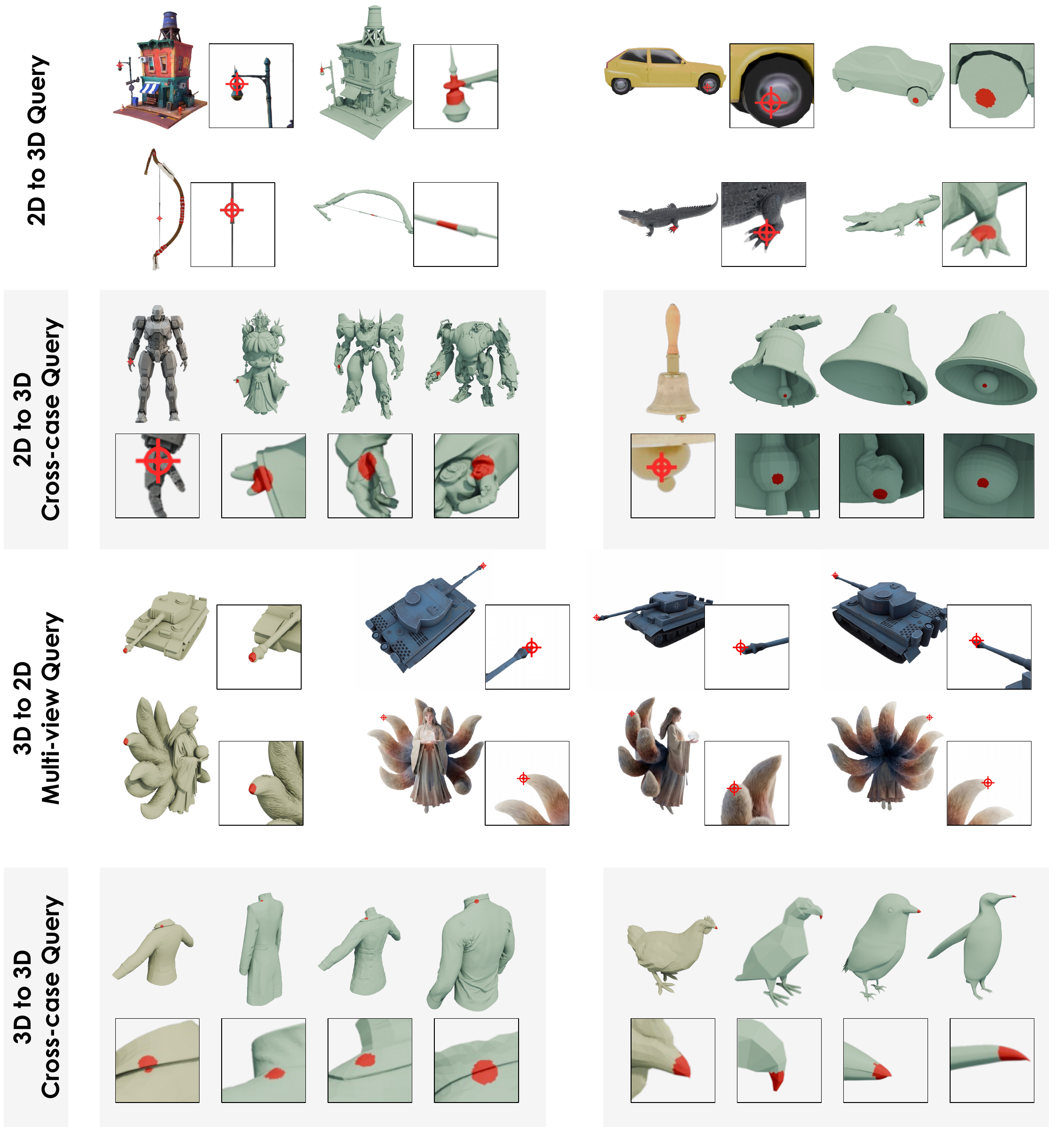}
    \caption{\textbf{Fine-grained local alignment.} The shared token space establishes accurate dense correspondences across diverse cross-modal and intra-modal scenarios.}
    \label{fig:local_vis}
    \vspace{-6mm}
\end{figure}

To aggregate the shared tokens into instance-level representations, the global branch extracts a single descriptor per modality. For the 2D modality, we first pool the shared 2D tokens within each view to form preliminary view tokens $\tilde{r}_s$. To inject rich semantics without compromising the established geometric grounding, we enrich these view tokens using the DPT context $c_s$ and the DINO class token $d_s$~\cite{oquab2024dinov} from the VGGT backbone:
\[
r_s = \tilde{r}_s + \lambda_c\, U c_s + \lambda_d\, \gamma_s \odot W_d d_s, \qquad \gamma_s = \sigma\!\bigl(\mathrm{MLP}([\tilde{r}_s;\, W_d d_s])\bigr),
\]
with $\lambda_c = 0.5$ and $\lambda_d = 1.0$. By applying this gated DINO fusion strictly after local evidence is pooled, we ensure the DINO signal acts as semantic enrichment rather than a shortcut that bypasses local alignment. A residual MLP then refines these view tokens before a second stage pools them across all valid views to produce the normalized 2D global descriptor $g^{2\mathrm{D}}$.

For the 3D modality, the shared 3D tokens first pass through a self-attention~\cite{vaswani2017attention} layer and a feed-forward network. This step enables spatially distant parts of the shape to exchange context. The contextualized 3D tokens are subsequently pooled, projected, and normalized to yield the 3D global descriptor $g^{3\mathrm{D}}$.

To align these global representations, we apply a symmetric InfoNCE cross-modal contrastive loss over the distributed batch of size $B$:
\[
\mathcal{L}_{\mathrm{global}} = -\frac{1}{2B} \sum_{i=1}^{B} \Biggl[ \log \frac{\exp(g_i^{2\mathrm{D}} \cdot g_i^{3\mathrm{D}} / \tau_g)}{\sum_{j=1}^{B} \exp(g_i^{2\mathrm{D}} \cdot g_j^{3\mathrm{D}} / \tau_g)} + \log \frac{\exp(g_i^{3\mathrm{D}} \cdot g_i^{2\mathrm{D}} / \tau_g)}{\sum_{j=1}^{B} \exp(g_i^{3\mathrm{D}} \cdot g_j^{2\mathrm{D}} / \tau_g)} \Biggr]
\],
where $\tau_g$ is the contrastive temperature, and $g_i$ denotes the global descriptor for the $i$-th sample in the batch. We further regularize the 2D descriptor against missing views via a subset-consistency loss $\mathcal{L}_{\mathrm{sub}}$, requiring random view subsets to match the full-view descriptor.

Finally, inspired by relation distillation~\cite{wang2024multi}, we introduce a semantic distillation loss $\mathcal{L}_{\mathrm{sd}}$ to transfer semantic knowledge from the 2D teacher. Instead of regressing the teacher features directly, we distill their relational structure. Given the mean DINO class token $t_i$ (where $t_i$ is the average of the class tokens $d_s$ over valid views of sample $i$), we compute a teacher similarity matrix and align it with the student's cross-modal similarity:
\[
\mathcal{L}_{\mathrm{sd}} = \mathrm{KL}\!\Bigl(\mathrm{softmax}\bigl(\frac{T T^\top}{\tau_d}\bigr) \;\Big\|\; \mathrm{softmax}\bigl(\frac{G^{2\mathrm{D}} (G^{3\mathrm{D}})^\top}{\tau_d}\bigr)\Bigr),
\]
where $\tau_d$ is the distillation temperature. This formulation encourages the joint feature space to inherit the semantic neighborhood structure of the teacher while remaining inherently cross-modal.

\begin{figure}[t]
\centering
\includegraphics[width=\linewidth]{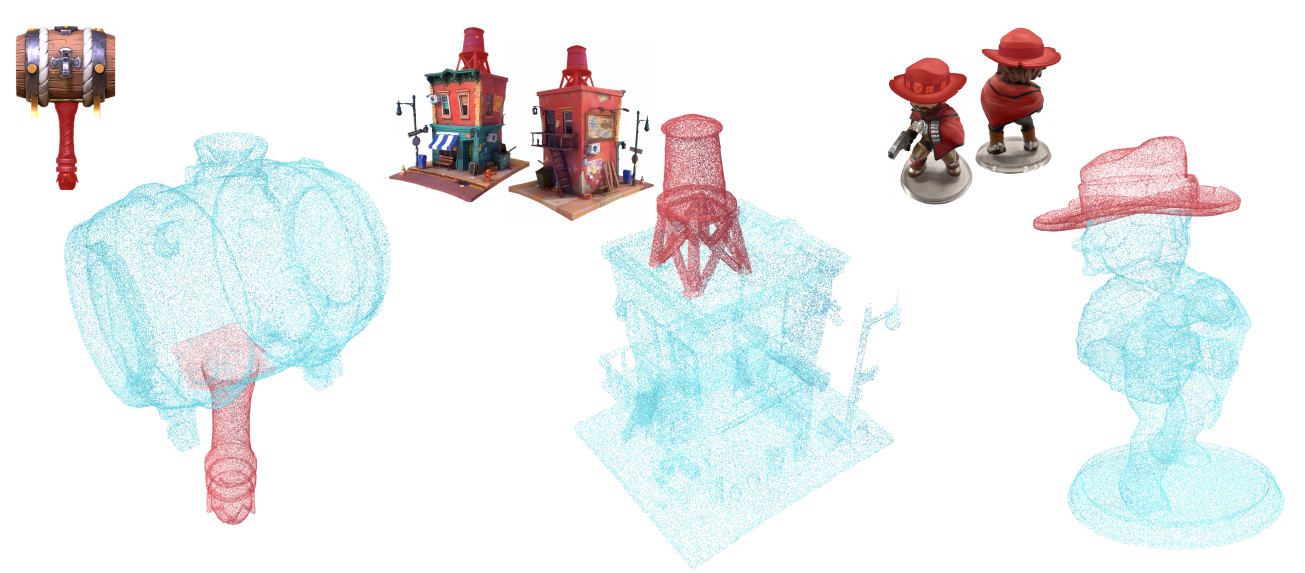}
\caption{\textbf{2D-to-3D part transfer.} 2D regions selected via SAM is mapped onto the 3D point clouds using our local descriptor space.}
\label{fig:part_vis}
\vspace{-2mm}
\end{figure}

\begin{table}[t]
\centering
\caption{\textbf{Image-to-shape retrieval.} Category-level Recall@$k$ and MRR on our ModelNet40 benchmark. Baselines use a single random view ($S{=}1$); our model is evaluated under both single-view and multi-view settings.}
\label{tab:retrieval}
\small
\setlength{\tabcolsep}{6pt}
\begin{tabular}{llccccc}
\toprule
Method & Input views & R@1 & R@2 & R@3 & R@5 & MRR \\
\midrule
Uni3D-g~\cite{ICLR2024_cc2c03ce}       & $S{=}1$ & 32.60 & 41.68 & 45.63 & 53.43 & 42.70 \\
OpenShape~\cite{NEURIPS2023_8c7304e7}   & $S{=}1$ & \textbf{36.44} & \textbf{47.50} & \textbf{52.74} & \underline{59.02} & \textbf{48.19} \\
ULIP-2~\cite{Xue_2024_CVPR}            & $S{=}1$ & 33.88 & 42.96 & 48.54 & 54.60 & 43.98 \\
\textbf{\modelname(ours)}  & $S{=}1$       & \underline{35.04} & \underline{45.17} & \underline{52.50} & \textbf{62.28} & \underline{47.17} \\
\midrule
\textbf{\modelname(ours)}  & $S{=}4$, ortho       & 37.60 & 49.94 & \textbf{56.23} & 63.10 & 49.67 \\
\textbf{\modelname(ours)}  & $S{=}4$, random      & \textbf{39.12} & \textbf{50.64} & 55.88 & \textbf{63.21} & \textbf{50.69} \\
\bottomrule
\end{tabular}
\vspace{-3mm}
\end{table}

\subsection{Progressive training}
\label{subsec:progressive_training}
To stably couple the dense and global objectives, we train the model in three progressive stages.
In Stage I, we disable the global branch and train only the shared encoders, local heads, and the 3D VAE encoder using $\mathcal{L}_{\mathrm{local}}$ without hard-negative mining. This establishes the foundational pixel-to-point geometric alignment.
In Stage II, we enable the global branch to integrate semantic representations. The model is jointly optimized, and the full objective combines $\mathcal{L}_{\mathrm{local}}$, $\mathcal{L}_{\mathrm{global}}$, $\mathcal{L}_{\mathrm{sub}}$, and $\mathcal{L}_{\mathrm{sd}}$. To preserve the correspondence structure learned in Stage I, we apply a reduced learning rate to the local modules, while introducing hard-negative mining to the local loss.
In Stage III, we scale up the 2D input resolution and apply strengthened hard-negative mining. All modules are jointly fine-tuned to sharpen overall feature discriminability.
Concrete hyperparameters, precise loss formulations, and learning-rate schedules for each stage are detailed in Table~\ref{tab:train-config-per-stage} (Appendix~\ref{sec:app-per-stage-config}).

\begin{figure}[t]
\centering
\includegraphics[width=\linewidth]{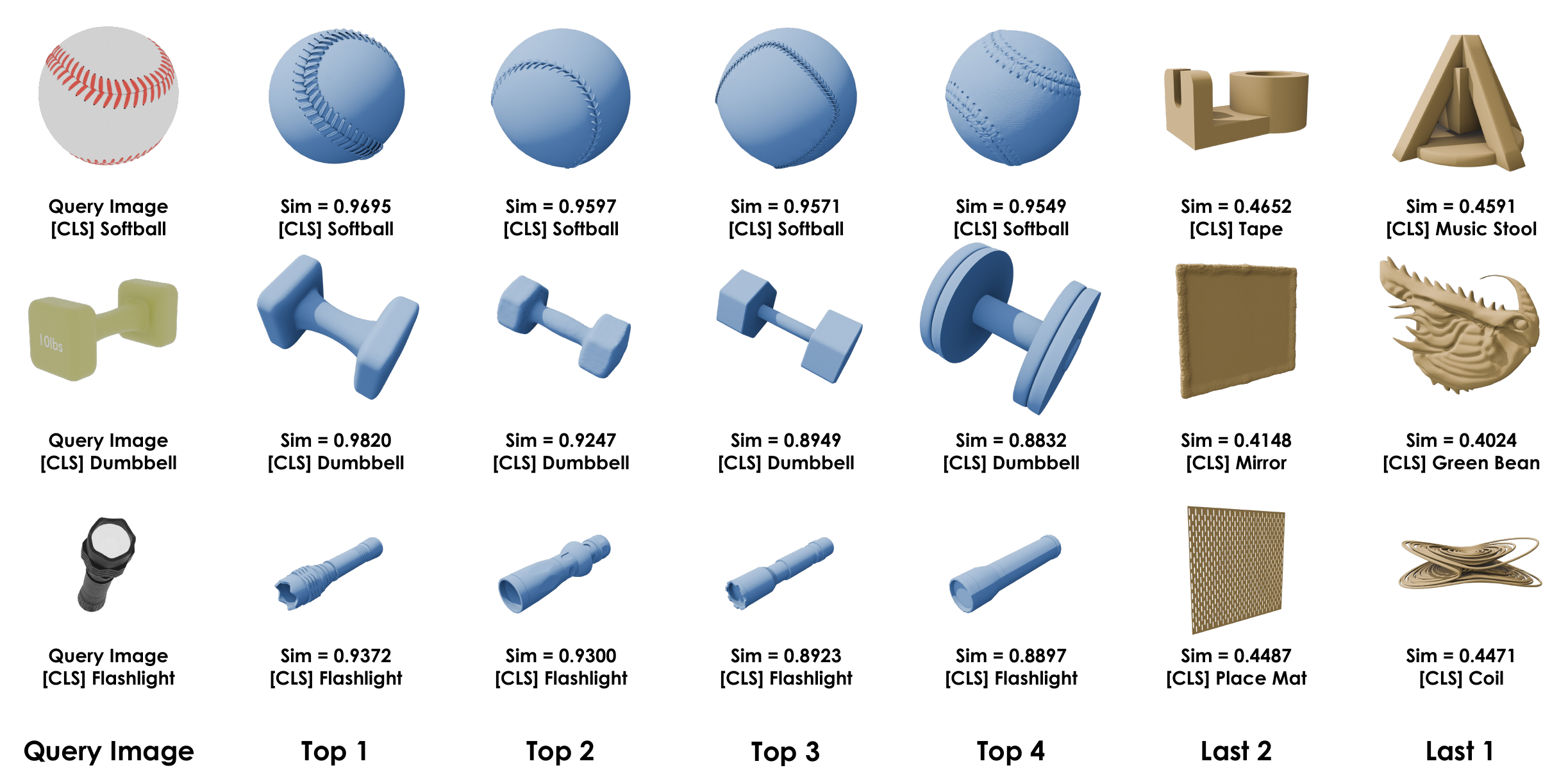}
\caption{\textbf{Image-to-shape retrieval.} Each row shows a query image (left), the top-4 retrieved 3D shapes ranked by global descriptor similarity, and the bottom-2 shapes for contrast. Similarity scores and predicted category labels are annotated below each result.}
\label{fig:global_2d3d}
\vspace{-5mm}
\end{figure}

\begin{figure}[t]
\centering
\includegraphics[width=\linewidth]{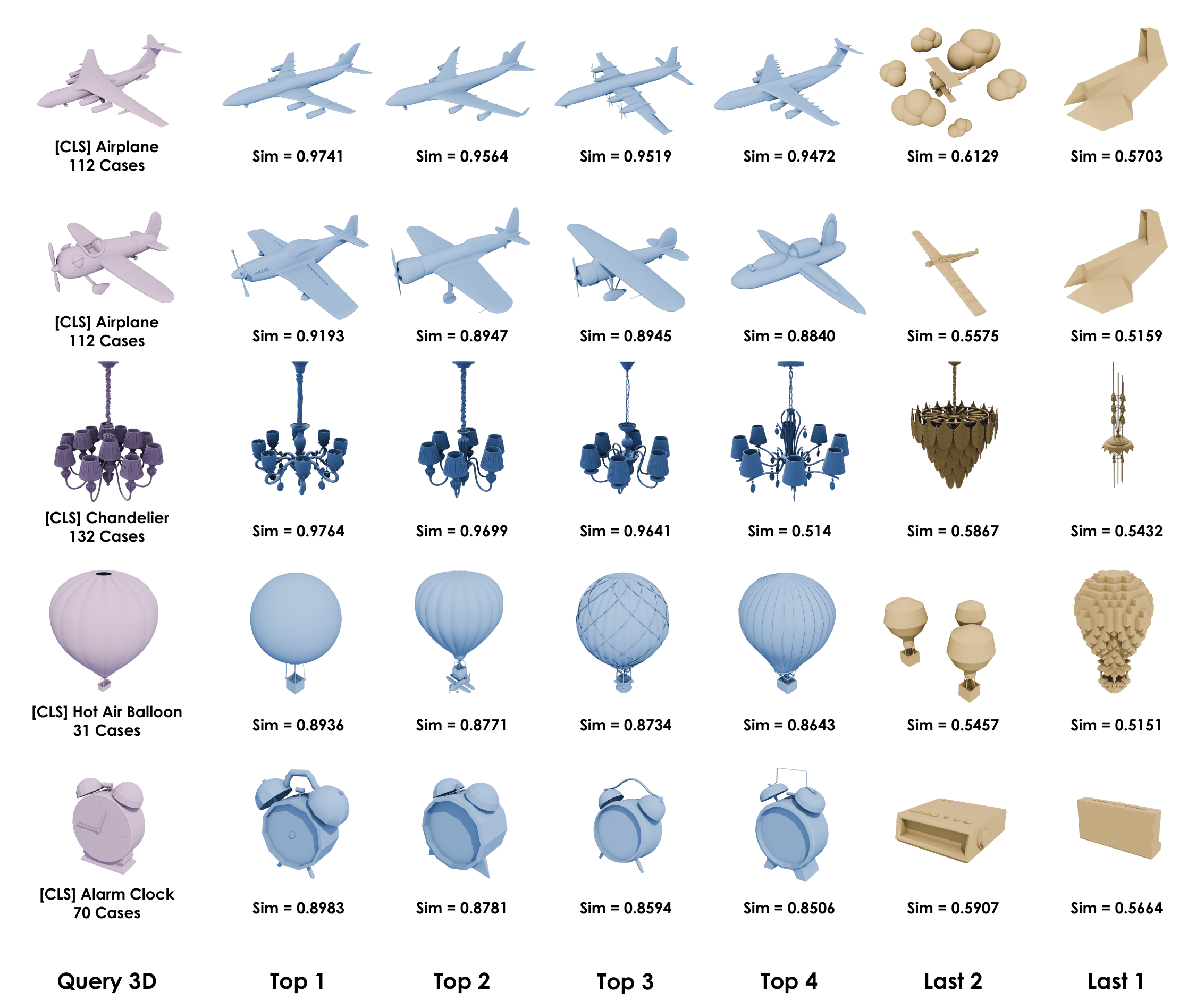}
\caption{\textbf{3D-to-3D shape retrieval.} Each row shows a query shape (left), the top-4 retrieved shapes, and the bottom-2 shapes using only the 3D global descriptor, without any 2D input. Retrieved shapes share fine-grained geometric and topological structure beyond basic category labels.}
\label{fig:global_3d3d}
\vspace{-5mm}
\end{figure}

\section{Experiments}
\label{sec:experiments}


\textbf{Training data.} We train \modelname on a large-scale 2D-3D dataset comprising approximately $650{,}000$ objects sourced from Objaverse~\cite{deitke2023objaverse,deitke2023objaversexl}, including a portion of the Objaverse-LVIS dataset. For the 3D modality, we extract a surface point cloud of $20{,}480$ points with coordinates and normals for each object. For the 2D modality, we render 60 views (6 orthographic and 54 random) per object, pairing each view with a RGBA position map. During training, we uniformly sample $S \in \{1,2,3,4\}$ views per object.

\textbf{Evaluation benchmark.} To prevent data leakage from the training set's inclusion of Objaverse-LVIS, we construct a custom benchmark rather than relying on standard LVIS evaluation splits. This benchmark comprises $3{,}055$ objects covering the 40 ModelNet40~\cite{3DShapeNets} categories. Since the original ModelNet40 meshes lack the textures required for multi-view rendering, we source textured instances from a mixture of generated assets and public 3D collections. Each test object includes 10 rendered views and a paired surface point cloud. Per-class statistics are detailed in Appendix~\ref{sec:app-benchmark-stats}.

\textbf{Baselines and metrics.} For pixel-to-point correspondence, no prior work addresses this task at the object level. As a reference baseline, we repurpose VGGT~\cite{wang2025vggt} and input four orthographic views (VGGT requires a known reference frame anchored by the front view) and directly read its pointmap prediction at each query pixel. We evaluate both the original VGGT and VGGT-OBJ (fine-tuned on object data, proposed in~\cite{chang2026reconviagen}). We measure this spatial alignment using LocAcc@$k$ (Localization Accuracy), defined as the mean of $(1 - d^\star_k / d_{\mathrm{norm}}) \times 100\%$. Here, $d^\star_k$ is the minimum Euclidean distance between the ground-truth 3D coordinate and the centers of the top-$k$ retrieved 3D tokens, and $d_{\mathrm{norm}} = \sqrt{3}L$ normalizes the error by the 3D bounding box diagonal. For image-to-shape retrieval, we compare against three leading global alignment models: Uni3D-g~\cite{ICLR2024_cc2c03ce}, OpenShape~\cite{NEURIPS2023_8c7304e7}, and ULIP-2~\cite{Xue_2024_CVPR}, reporting standard Recall@$\{1,2,3,5\}$ and Mean Reciprocal Rank (MRR).

\subsection{Pixel-to-point correspondence}
\label{sec:exp-local}

To evaluate fine-grained local alignment, we randomly sample up to 20 foreground pixels per view and decode their ground-truth 3D coordinates using the corresponding position maps. For each sampled pixel, we retrieve the 3D token with the highest cosine similarity in the shared local descriptor space. Table~\ref{tab:local} reports the LocAcc performance under three view configurations. Our model achieves a LocAcc@1 above 75 across all settings, peaking at 77.37 given four random views.

Beyond quantitative metrics, Figure~\ref{fig:local_vis} demonstrates the versatility of our learned local descriptor space across diverse matching scenarios. In standard 2D-to-3D intra-object matching, a single query pixel accurately localizes the corresponding geometric region on the 3D mesh, confirming robust fine-grained spatial grounding.
This capability naturally extends to cross-case 2D-to-3D queries, where a pixel on one object successfully retrieves the semantically equivalent 3D region on an entirely different instance, indicating that the shared space captures strong category-level semantics rather than merely overfitting to exact shapes.
Furthermore, reverse 3D-to-2D queries show that a single 3D point correctly grounds back to corresponding pixels across diverse camera poses, verifying strict multi-view consistency.
Finally, pure 3D-to-3D cross-case queries reveal that the 3D tokens themselves acquire semantic part-level alignment, accurately matching corresponding structural parts across different shapes without relying on any 2D inputs during inference.

Our pixel-to-point alignment naturally extends from individual points to contiguous semantic parts. By coupling a SAM~\cite{ravi2025sam} with our local descriptor space, we can transfer an arbitrary 2D part mask onto the 3D mesh without any part-level supervision. As shown in Figure~\ref{fig:part_vis}, given a user-selected 2D region, our pipeline identifies the corresponding 3D surface region with high fidelity. This demonstrates that the dense alignment learned by \modelname is coherent enough to support region-level transfer, bridging pixel-level matching and part-level understanding. Details of the transfer pipeline are provided in Appendix~\ref{sec:app-part-pipeline}.

\subsection{Global retrieval}
\label{sec:exp-retrieval}

For category-level image-to-shape retrieval, we use the 2D global descriptor of the query image(s) to search against the 3D global descriptors of all test objects. A match is considered correct if the retrieved shape belongs to the same category as the query. Table~\ref{tab:retrieval} compares our performance against the global-only baselines. Across both single-view and multi-view configurations, \modelname achieves highly competitive retrieval accuracy alongside established models like OpenShape.

Qualitative results in Figure~\ref{fig:global_2d3d} further illustrate this capability. The top-ranked 3D shapes align with the 2D queries not only in broad category semantics but also in fine-grained structural details, demonstrating the benefit of grounding global representations on local geometric features.

Beyond cross-modal matching, our 3D global descriptors naturally support pure shape-to-shape retrieval without any 2D input. As shown in Figure~\ref{fig:global_3d3d}, the retrieved shapes share not only category semantics but also fine-grained geometric and topological similarities. This confirms that joint training with the local geometric branch gives the 3D encoder richer structural awareness than a purely semantic objective would.

\subsection{Ablation on progressive training}
\label{sec:exp-ablation}

To validate our three-stage progressive training strategy, we evaluate the checkpoint at the end of each stage on both tasks (Table~\ref{tab:ablation-stages}). Stage~I, trained with only the local objective, establishes strong pixel-to-point correspondence but provides no global retrieval capability. After Stage~II introduces the global branch, the model gains competitive retrieval performance while the local metrics are largely preserved or improved, confirming that the two objectives reinforce rather than conflict with each other. Stage~III further boosts both tasks through higher-resolution input and strengthened hard-negative mining. The monotonic improvement across all metrics validates the effectiveness of each stage. Full per-stage breakdowns are in Appendix~\ref{sec:app-per-stage-local} and~\ref{sec:app-per-stage-retrieval}.

\begin{table}[t]
\centering
\caption{\textbf{Progressive training ablation.} We evaluate the model at the end of each training stage. The improvement in both pixel-to-point localization and image-to-shape retrieval demonstrates that progressively introducing global aggregation and scaling resolution steadily enhances both geometric and semantic representations.}
\label{tab:ablation-stages}
\small
\setlength{\tabcolsep}{4pt}
\begin{tabular}{lccccc}
\toprule
 & \multicolumn{2}{c}{Local (pixel $\!\to\!$ point)} & \multicolumn{3}{c}{Global (image $\!\to\!$ shape)} \\
\cmidrule(lr){2-3}\cmidrule(lr){4-6}
Stage & LocAcc@1 & LocAcc@10 & R@1 & R@10 & MRR \\
\midrule
Stage~I        & 76.47 & 86.42 & -   & -   & - \\
Stage~II     & 76.99 & 87.35 & 36.79 & 71.94 & 47.91 \\
Stage~III & 77.37 & 87.38 & 39.12 & 74.16 & 50.69 \\
\bottomrule
\end{tabular}
\end{table}

\section{Conclusion and limitation}
\label{sec:conclusion}

We presented \modelname, a foundation model that unifies dense correspondence and global retrieval within a single shared token space. We adopt a geometry-aware 2D backbone as the alignment anchor and encode point clouds into discrete 3D tokens via a pretrained VAE. This design enables fine-grained 2D-3D alignment at the object level. To stabilize the joint optimization of dense and global objectives, we introduce a three-stage progressive training strategy that ensures the two tasks reinforce each other. Experiments confirm that \modelname achieves strong pixel-to-point localization while remaining competitive with leading global retrieval baselines. Furthermore, the originally non-semantic 3D encoder acquires discriminative semantic structure through cross-modal training alone, showing strong potential for a wide range of downstream tasks.

Our current local alignment operates at a compressed resolution on both sides: VGGT compresses images into patch-level features where each token covers a $14 \times 14$ pixel region, and the 3D VAE compresses point clouds into 1024 discrete tokens. Both compression steps discard sub-patch and sub-token spatial details, placing an inherent error on point-level localization accuracy. In addition, our global branch relies on DINO class tokens for semantic enrichment, which carry weaker category-level discriminability compared to CLIP features. This gap makes our single-view retrieval still trail OpenShape on several metrics.

\clearpage

\bibliographystyle{unsrt}
\bibliography{ref}

\clearpage
\appendix
%

\section{Appendix}

\subsection{Evaluation benchmark statistics}
\label{sec:app-benchmark-stats}

Table~\ref{tab:benchmark-per-class} reports the per-class object counts.

\begin{table}[h]
\centering
\caption{\textbf{Per-class object counts for the evaluation benchmark} We detail the distribution of the $3{,}055$ textured objects across 40 ModelNet40 categories.}
\label{tab:benchmark-per-class}
\small
\setlength{\tabcolsep}{4pt}
\begin{tabular}{lr@{\hspace{1.5em}}lr@{\hspace{1.5em}}lr@{\hspace{1.5em}}lr}
\toprule
Category & \# & Category & \# & Category & \# & Category & \# \\
\midrule
airplane   & 85 & cup        & 74 & laptop       & 83 & sofa     & 66 \\
bathtub    & 97 & curtain    & 89 & mantel       & 75 & stairs   & 45 \\
bed        & 29 & desk       & 59 & monitor      & 97 & stool    & 80 \\
bench      & 75 & door       & 40 & night\_stand & 98 & table    & 57 \\
bookshelf  & 70 & dresser    & 83 & person       & 86 & tent     & 78 \\
bottle     & 70 & flower\_pot & 69 & piano       & 98 & toilet   & 57 \\
bowl       & 79 & glass\_box & 54 & plant        & 74 & tv\_stand & 97 \\
car        & 75 & guitar     & 86 & radio        & 93 & vase     & 86 \\
chair      & 60 & keyboard   & 65 & range\_hood  & 84 & wardrobe & 84 \\
cone       & 85 & lamp       & 95 & sink         & 93 & xbox     & 85 \\
\bottomrule
\end{tabular}
\end{table}

\subsection{Per-stage training configurations}
\label{sec:app-per-stage-config}

Table~\ref{tab:train-config-per-stage} lists the per-stage hyperparameters. Entries left unchanged across all three stages (MLP widths, layer norms, $[1,4]$ view sampling, $N{=}20{,}480$ surface points, $512$ FPS query budget, 2D Gaussian-blur augmentation, DeepSpeed~ZeRO-2 with bf16, AdamW with weight decay $0.01$ and gradient clipping at $1.0$) are omitted for brevity.

\begin{table}[t]
\centering
\caption{\textbf{Per-stage training configurations.} Detailed hyperparameter evolution across the three progressive training stages.}
\label{tab:train-config-per-stage}
\small
\begin{tabular}{lccc}
\toprule
 & Stage~I & Stage~II & Stage~III \\
\midrule
\multicolumn{4}{l}{\textit{Branch / freezing schedule}} \\
\quad VGGT backbone (frozen)     & \ding{51} & \ding{51} & \ding{51} \\
\quad 3D VAE encoder trainable   & \ding{51} & \ding{51} & \ding{51} \\
\quad Shared encoders trainable  & \ding{51} & \ding{51} (low LR) & \ding{51} (low LR) \\
\quad Local heads trainable      & \ding{51} & \ding{51} (low LR) & \ding{51} \\
\quad Enable global branch       & \ding{55} & \ding{51} & \ding{51} \\
\quad Global heads trainable     & \ding{55} & \ding{51} & \ding{51} \\
\midrule
\multicolumn{4}{l}{\textit{Architecture toggles (global branch)}} \\
\quad DINO cls fusion (2D)     & \ding{55} & \ding{51} & \ding{51} \\
\midrule
\multicolumn{4}{l}{\textit{2D input resolution}} \\
\quad VGGT input size          & $518{\times}518$ & $518{\times}518$ & $1022{\times}1022$ \\
\quad Patch grid               & $37{\times}37$   & $37{\times}37$   & $73{\times}73$ \\
\midrule
\multicolumn{4}{l}{\textit{Loss weights}} \\
\quad $\lambda_{\mathrm{local}}$  & 1.0   & 0.25 & 0.50 \\
\quad $\lambda_{\mathrm{global}}$ & 0.0   & 1.0  & 0.80 \\
\quad $\lambda_{\mathrm{sub}}$    & 0.0   & 0.05 & 0.05 \\
\quad $\lambda_{\mathrm{sd}}$     & n/a   & 0.10 & 0.20 \\
\midrule
\multicolumn{4}{l}{\textit{Local loss details}} \\
\quad Loss mode                 & soft-only    & +hard-neg    & +hard-neg \\
\quad Hard-negative top-$k$     & n/a          & 64           & 96 \\
\quad Hard-negative weight      & n/a          & 0.25         & 0.50 \\
\quad Spatial negative radius $\delta$ & 0.020 & 0.020        & 0.015 \\
\midrule
\multicolumn{4}{l}{\textit{Distillation temperature}} \\
\quad $\tau_d$                    & n/a   & 0.07 & 0.05 \\
\midrule
\multicolumn{4}{l}{\textit{Optimization}} \\
\quad Base learning rate         & $1{\times}10^{-4}$ & $6{\times}10^{-5}$ & $3{\times}10^{-5}$ \\
\quad Shared LR scale            & 1.0   & 0.10 & 0.05 \\
\quad Local LR scale             & 1.0   & 0.30 & 0.50 \\
\quad Global LR scale            & n/a   & 1.0  & 1.0  \\
\quad Per-GPU batch size         & 30    & 25   & 7    \\
\quad Training length (epochs)   & 5     & 3    & 3    \\
\bottomrule
\end{tabular}
\end{table}

\subsection{Per-stage pixel-to-point local correspondence}
\label{sec:app-per-stage-local}

Table~\ref{tab:local-per-stage} reports LocAcc at each training stage under all view configurations.

\begin{table}[t]
\centering
\caption{\textbf{Per-stage pixel-to-point local correspondence.} LocAcc improves steadily across stages, with Stage~III benefiting from higher resolution and stronger hard-negative mining.}
\label{tab:local-per-stage}
\small
\begin{tabular}{llccccc}
\toprule
Stage & Input views & LocAcc@1 & LocAcc@2 & LocAcc@3 & LocAcc@5 & LocAcc@10 \\
\midrule
\multirow{3}{*}{Stage~I}
 & $S{=}1$ (random)       & 76.40 & 79.55 & 81.37 & 83.55 & 86.32 \\
 & $S{=}4$ (random4)      & 76.47 & 79.62 & 81.46 & 83.69 & 86.42 \\
 & $S{=}4$ (ortho4)       & 75.58 & 78.90 & 80.86 & 83.23 & 86.14 \\
\midrule
\multirow{3}{*}{Stage~II}
 & $S{=}1$ (random)       & 76.75 & 79.74 & 81.60 & 83.95 & 87.01 \\
 & $S{=}4$ (random4)      & 76.99 & 80.00 & 81.88 & 84.23 & 87.35 \\
 & $S{=}4$ (ortho4)       & 76.01 & 79.19 & 81.17 & 83.64 & 86.88 \\
\midrule
\multirow{3}{*}{Stage~III}
 & $S{=}1$ (random)       & 77.43 & 80.24 & 82.02 & 84.23 & 87.14 \\
 & $S{=}4$ (random4)      & 77.37 & 80.26 & 82.07 & 84.37 & 87.38 \\
 & $S{=}4$ (ortho4)       & 76.37 & 79.49 & 81.42 & 83.85 & 87.00 \\
\bottomrule
\end{tabular}
\end{table}

\subsection{Per-stage image-to-shape retrieval}
\label{sec:app-per-stage-retrieval}

Table~\ref{tab:retrieval-per-stage} reports retrieval metrics at each training stage under all view configurations.

\begin{table}[t]
\centering
\caption{\textbf{Per-stage image-to-shape retrieval.} Retrieval performance following the activation of the global branch in Stage~II. The subsequent joint optimization in Stage~III further enhances Recall and MRR.}
\label{tab:retrieval-per-stage}
\small
\setlength{\tabcolsep}{4pt}
\begin{tabular}{llcccccc}
\toprule
Stage & Input views & R@1 & R@2 & R@3 & R@5 & R@10 & MRR \\
\midrule
\multirow{3}{*}{Stage~II}
 & $S{=}1$ (random)       & 33.76 & 46.22 & 53.20 & 60.19 & 71.25 & 45.44 \\
 & $S{=}4$ (ortho4)       & 35.27 & 46.92 & 54.13 & 59.37 & 67.05 & 45.77 \\
 & $S{=}4$ (random4)      & 36.79 & 48.54 & 55.88 & 63.33 & 71.94 & 47.91 \\
\midrule
\multirow{3}{*}{Stage~III}
 & $S{=}1$ (random)       & 35.04 & 45.17 & 52.50 & 62.28 & 71.83 & 47.17 \\
 & $S{=}4$ (ortho4)       & 37.60 & 49.94 & 56.23 & 63.10 & 71.59 & 49.67 \\
 & $S{=}4$ (random4)      & 39.12 & 50.64 & 55.88 & 63.21 & 74.16 & 50.69 \\
\bottomrule
\end{tabular}
\end{table}

\subsection{Additional pixel-to-point correspondence visualizations}
\label{sec:app-more-local-vis}

\begin{figure}[t]
\centering
\includegraphics[width=\linewidth]{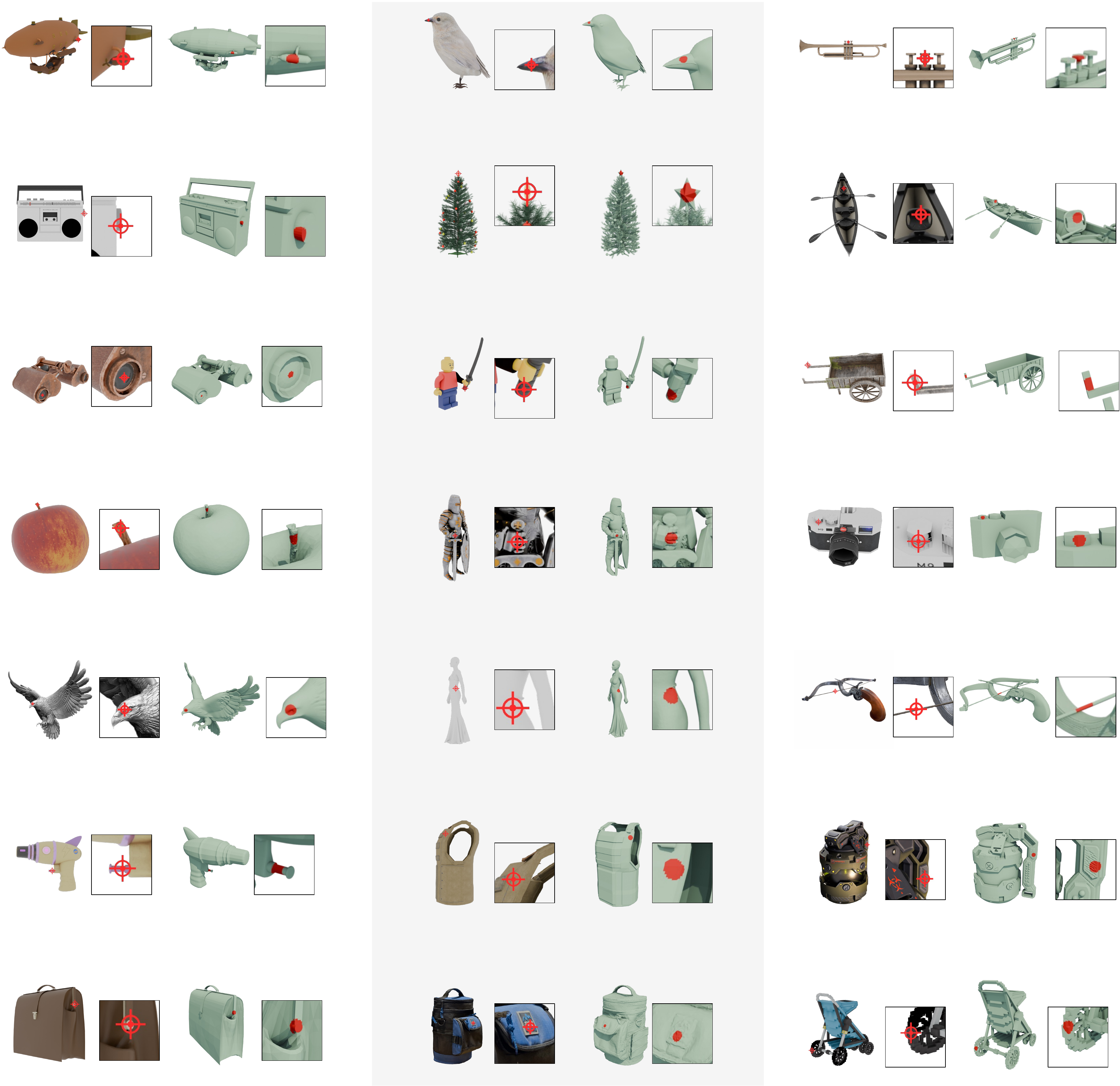}
\caption{\textbf{Additional pixel-to-point correspondence visualizations.} More examples of pixel-to-point matching, complementing Figure~\ref{fig:local_vis}.}
\label{fig:local_vis_appendix}
\end{figure}

Figure~\ref{fig:local_vis_appendix} shows additional examples complementing Figure~\ref{fig:local_vis}.

\subsection{2D part to 3D part transfer pipeline}
\label{sec:app-part-pipeline}

Given an input image, we first apply SAM~\cite{ravi2025sam} in automatic mode to produce a set of non-overlapping part masks, and select the target part as the smallest mask containing a user-specified click point. The selected pixel-level mask is then mapped onto the VGGT patch grid by marking a patch as active whenever the mask covers a sufficient fraction of its area, and the local 2D descriptors at all active patches are collected. Each of these 2D descriptors is matched against the full set of 3D token descriptors via cosine similarity, keeping only the top-1 3D token per patch and deduplicating tokens matched by multiple patches by retaining the highest-similarity assignment. To ensure spatial coherence, we discard matches with low similarity scores and apply DBSCAN clustering on the remaining 3D coordinates to keep only the dominant spatial cluster. Finally, the filtered 3D token centers are projected onto the mesh surface to identify nearby seed faces, from which a breadth-first-search flood fill along the face-adjacency graph propagates the region label to topologically connected faces; only the largest connected component is retained, yielding the final 3D part region. The entire pipeline operates at inference time and requires no part-level training or annotations.

\subsection{Additional shape-to-shape retrieval visualizations}
\label{sec:app-more-global-3d3d-vis}

\begin{figure}[t]
\centering
\includegraphics[width=\linewidth]{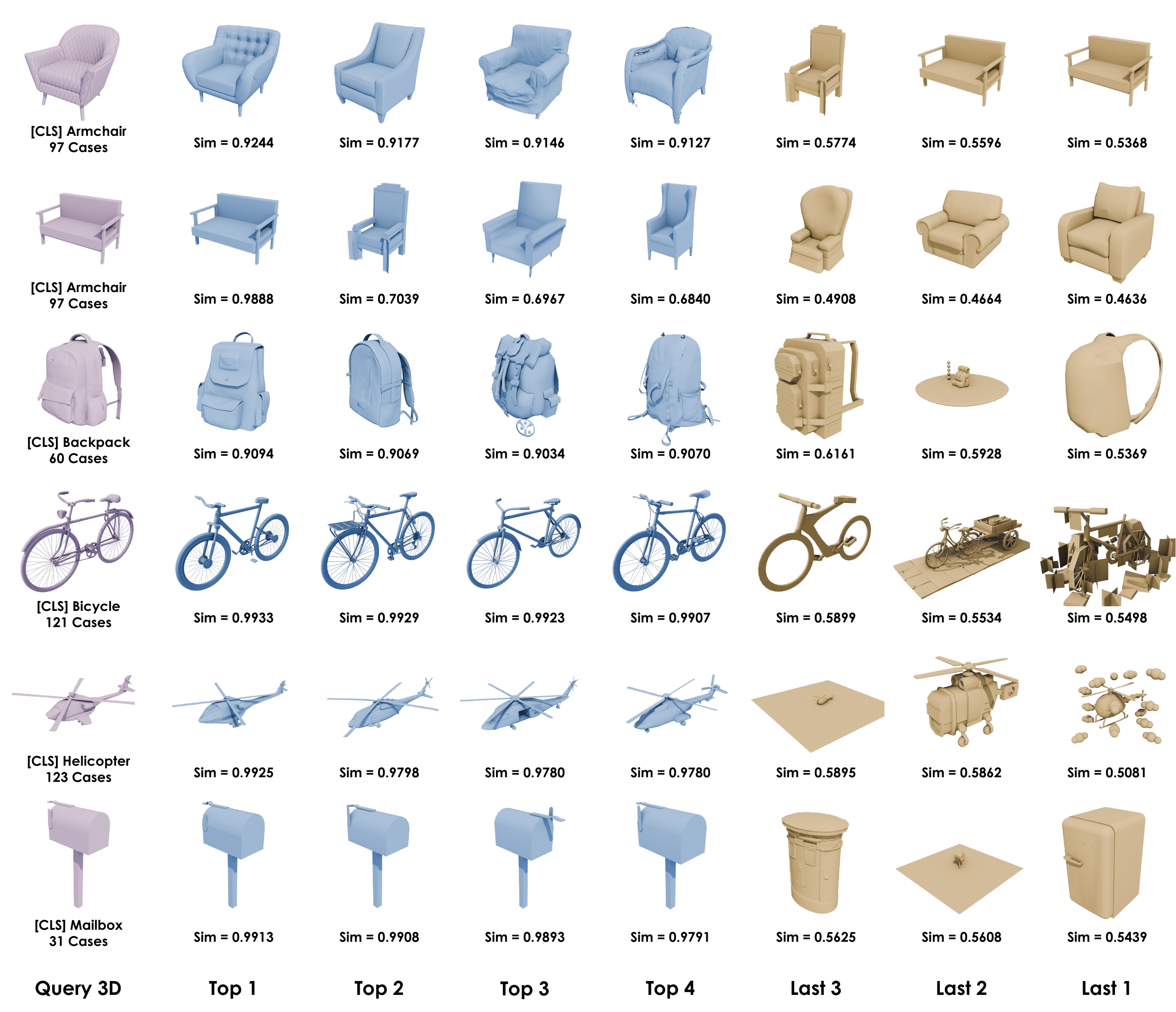}
\caption{\textbf{Additional 3D-to-3D shape retrieval visualizations.} More query-retrieval pairs showing that the 3D global descriptor captures fine-grained geometric similarity.}
\label{fig:global_3d3d_appendix}
\end{figure}

Figure~\ref{fig:global_3d3d_appendix} shows additional examples complementing Figure~\ref{fig:global_3d3d}.

\end{document}